\pdfoutput=1
\documentclass[11pt]{article}

\usepackage[]{acl}

\usepackage{times}
\usepackage{latexsym}

\usepackage{times}
\usepackage{latexsym}
\usepackage{subcaption}
\usepackage{caption}
\usepackage{graphicx}
\usepackage{multirow}
\usepackage{booktabs}
\usepackage{colortbl}
\usepackage{xcolor}
\usepackage{amsmath}
\usepackage{amssymb}
\usepackage{amsfonts}
\usepackage{algorithm}
\usepackage{algorithmic}
\usepackage{float}
\usepackage{bbm}
\usepackage{kotex}
\usepackage{mathtools}
\usepackage{booktabs}
\usepackage[utf8]{inputenc}
\usepackage{enumitem}
\usepackage{amsthm}

\usepackage[T1]{fontenc}
\usepackage[utf8]{inputenc}

\usepackage{microtype}

\usepackage{booktabs}
\usepackage{multirow}
\newcommand\modelname{SQuID}
\let\SUP\textsuperscript

\newcommand\jiho[1]{\textcolor{black}{{#1}}}
\newcommand\yeon[1]{\textcolor{black}{{#1}}}

\title{Two-Step Question Retrieval for Open-Domain QA}
\author{Yeon Seonwoo\SUP{$\dagger$*}{\normalfont ,} {\bf Juhee Son\SUP{$\dagger$*}{\normalfont ,}} {\bf Jiho Jin\SUP{$\dagger$}{\normalfont ,}}\\ {\bf Sang-Woo Lee\SUP{$\ddagger \mathsection$},} {\bf Ji-Hoon Kim\SUP{$\ddagger \mathsection$}{\normalfont ,}} {\bf Jung-Woo Ha\SUP{$\ddagger \mathsection$}{\normalfont ,}}\\ {\bf Alice Oh\SUP{$\dagger$}}\\
  \SUP{$\dagger$}KAIST,
  \SUP{$\ddagger$}NAVER AI Lab, \SUP{$\mathsection$}NAVER CLOVA\\
  {\tt \{yeon.seonwoo,sjh5665,jinjh0123\}@kaist.ac.kr}\\
  {\tt\{sang.woo.lee,genesis.kim,jungwoo.ha\}@navercorp.com}\\
  {\tt alice.oh@kaist.edu}\\
}

\begin{document}
\maketitle
\def\thefootnote{*}\footnotetext{These authors contributed equally.}
\renewcommand{\thefootnote}{\arabic{footnote}}
\begin{abstract}
The retriever-reader pipeline has shown promising performance in open-domain QA but suffers from a very slow inference speed.
Recently proposed question retrieval models tackle this problem by indexing question-answer pairs and searching for similar questions. These models have shown a significant increase in inference speed, but
at the cost of lower QA performance compared to the retriever-reader models. 
This paper proposes a two-step question retrieval model, \textbf{\modelname{}} (\textbf{S}equential \textbf{Qu}estion-\textbf{I}ndexed \textbf{D}ense retrieval) and distant supervision for training.
\modelname{} uses two bi-encoders for question retrieval.
The first-step retriever selects top-k similar questions, and the second-step retriever finds the most similar question from the top-k questions.
We evaluate the performance and the computational efficiency of \modelname{}.
The results show that \modelname{} significantly increases the performance of existing question retrieval models with a negligible loss on inference speed.\footnote{The implementation of \modelname{} has been released at \url{https://github.com/yeonsw/SQuID.git}}
\end{abstract}

\section{Introduction}
Retriever-reader models in open-domain QA require a long time for inference \cite{izacard2021leveraging, NEURIPS2020_6b493230, sachan2021end, mao-etal-2021-generation, karpukhin2020dense}. This has been identified as a bottleneck in building real-time QA systems, and question retrieval and phrase-indexed QA have been proposed to resolve this problem \cite{seo2018phrase, seo2019real, lee2020contextualized, lee-etal-2021-learning-dense, lee2021phrase, lewis2021question, lewis2021paq}.
These approaches directly search the answer of the input question from the corpus without conducting additional machine reading steps which are computationally inefficient.
In phrase-indexed QA, retrievers pre-index all phrases in the corpus and find the most similar phrase to the input question.
In question retrieval, synthetic question-answer pairs are pre-indexed and referenced by retrievers \cite{du2017learning, duan2017question, fabbri2020template, lewis2020bart}.

\begin{figure}
    \centering
    \includegraphics[width=0.45\textwidth]{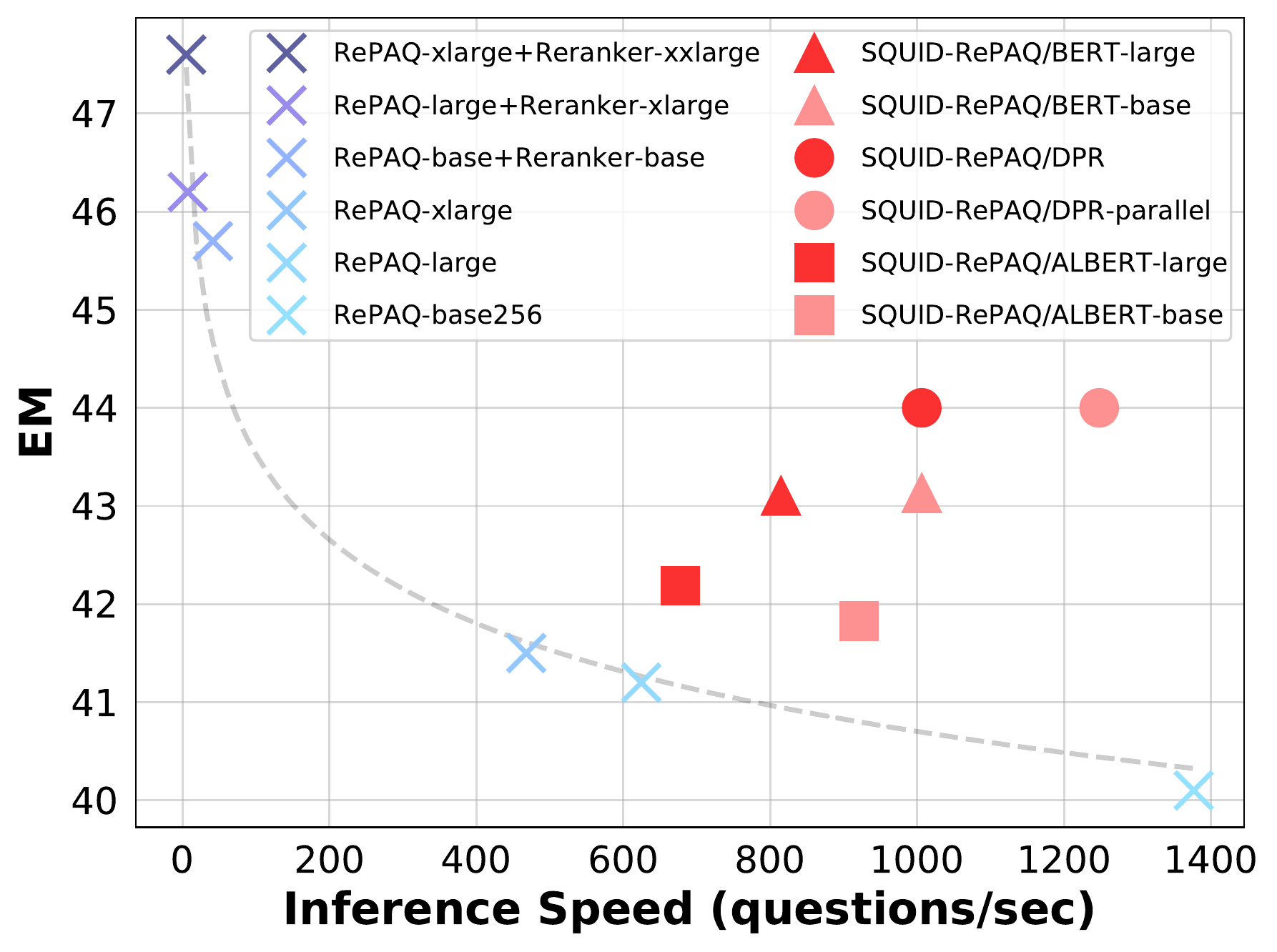}
    \caption{
    Trade-off relation between the open-domain QA performance and the inference time of existing question retrieval models (blue dots) and \modelname{} (red dots) on NaturalQuestions (NQ).
    The x-axis represents the inference speed and the y-axis represents the QA performance.
    }
    \label{fig:comp_eff}
    \vspace{-1.5em}
\end{figure}

Although recent question retrieval models significantly increase the inference speed, this improvement accompanies QA performance degradation.
Several approaches have been applied to question retrieval models to overcome the performance degradation, such as adopting the cross-encoder \cite{mao-etal-2021-reader, xiong2020answering} for re-ranking and increasing the model size \cite{lewis2021paq}.
However, these approaches cause a significant loss of computational efficiency.
Figure \ref{fig:comp_eff} shows the trade-off between the open-domain QA performance and the inference speed of question retrieval models.

We propose \textbf{\modelname{}} (\textbf{S}equential \textbf{Qu}estion-\textbf{I}ndexed \textbf{D}ense retrieval) which significantly improves QA performance without losing computational efficiency.
Our work follows previous work on neural re-ranking methods, which use a cross-encoder to re-rank the top-k passages retrieved from the first-step retriever \cite{lewis2021paq, xiong2020answering}.
Re-ranking methods have improved retrieval performance but require huge computation costs due to the cross-encoder architecture.
We use an additional bi-encoder retriever in \modelname{} instead of the cross-encoder to prevent loss on computational efficiency.
We also provide distant supervision methods for training the additional retriever in the absence of training data for question retrievers.

We evaluate \modelname{} on NaturalQuestions (NQ) \cite{kwiatkowski2019natural} and TriviaQA \cite{joshi2017triviaqa}.
We conduct three types of experiments: open-domain QA, computational efficiency evaluation, and analysis on distant supervision methods for training the second-step retriever.
Experimental results show that \modelname{} significantly outperforms the state-of-the-art question retrieval model by 4.0\%p on NQ and 6.1\%p on TriviaQA without losing computational efficiency.
Our main contribution is in proposing a sequential question retriever model that successfully improves both QA performance and inference speed, thereby making a meaningful step toward developing real-time open-domain QA systems.
\section{Related Work}
The research problem of reducing the computational cost of open-domain QA has received much attention recently.
The main bottleneck of a retriever-reader model is the machine reading step, and \newcite{seo2018phrase, seo2019real, lee-etal-2021-learning-dense} propose phrase-indexed QA, which directly retrieves the answer from the corpus without the machine reading step.
These models pre-compute the context of phrases in a corpus and conduct lexical and semantic similarity searches between the given question and the context of phrases \cite{zhao-etal-2021-sparta, yamada2021bpr}.
Most related to our work are the question retrieval models with question-generation models to build question-answer pairs and conduct a similarity search between the input question and the pre-indexed questions \cite{lewis2021question, lewis2021paq}. These models significantly reduce the computational cost but results in lower performance.
Our work provides an efficient question retrieval pipeline with distant supervision methods for training, while previous question retrieval models focus on the indexing methods with less attention on the retrieval pipeline.

\section{Method}

\begin{figure*}[ht]
    \begin{center}
        \begin{subfigure}{.53\textwidth}
            \centering
            \includegraphics[width=1.\linewidth]{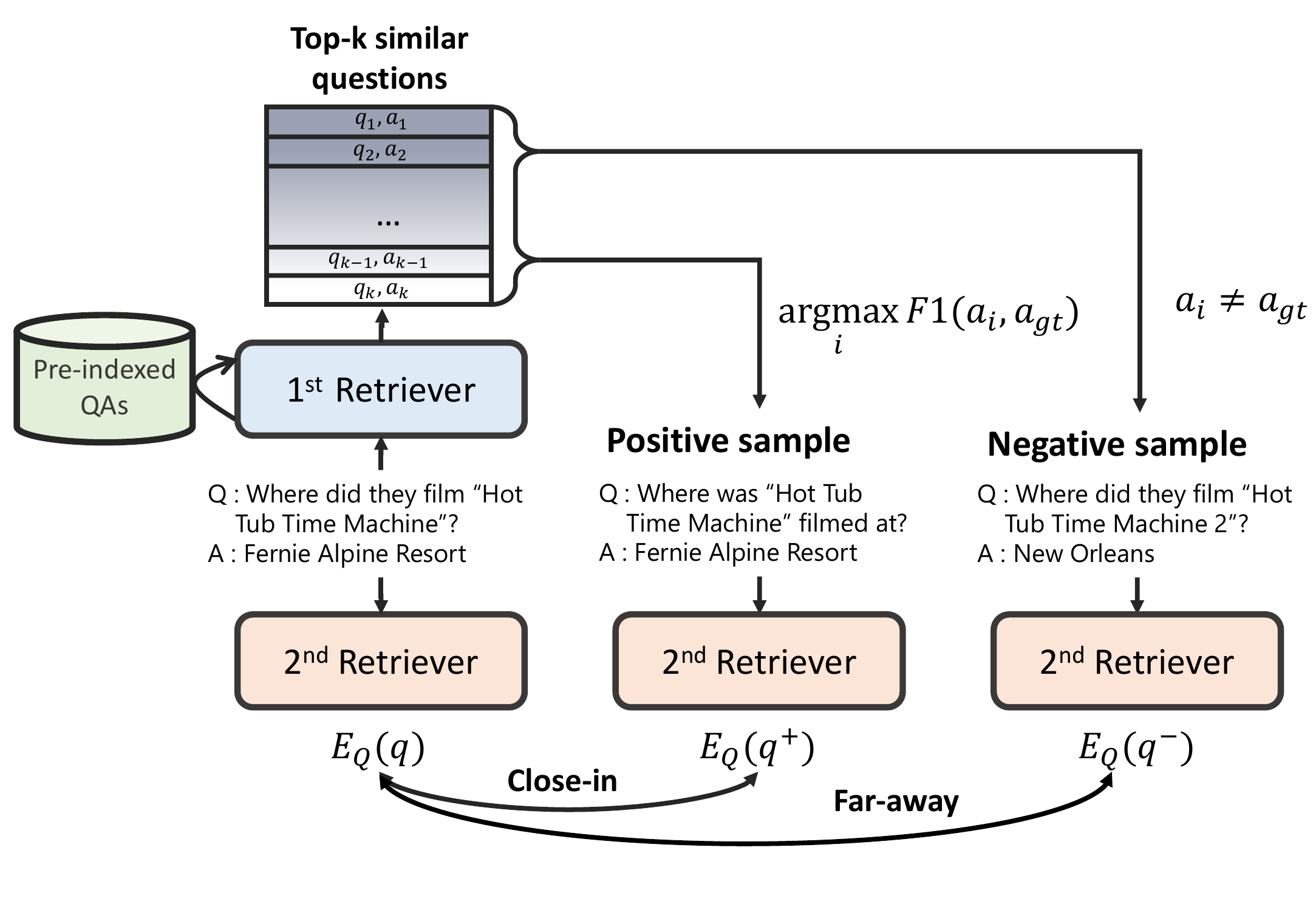}
            \vspace{-2.0em}
            \caption{Training procedure}
            \label{model_training}
        \end{subfigure}\hfill%
        \hspace{.02\textwidth}
        \begin{subfigure}{.35\textwidth}
            \centering
            \includegraphics[width=1.\linewidth]{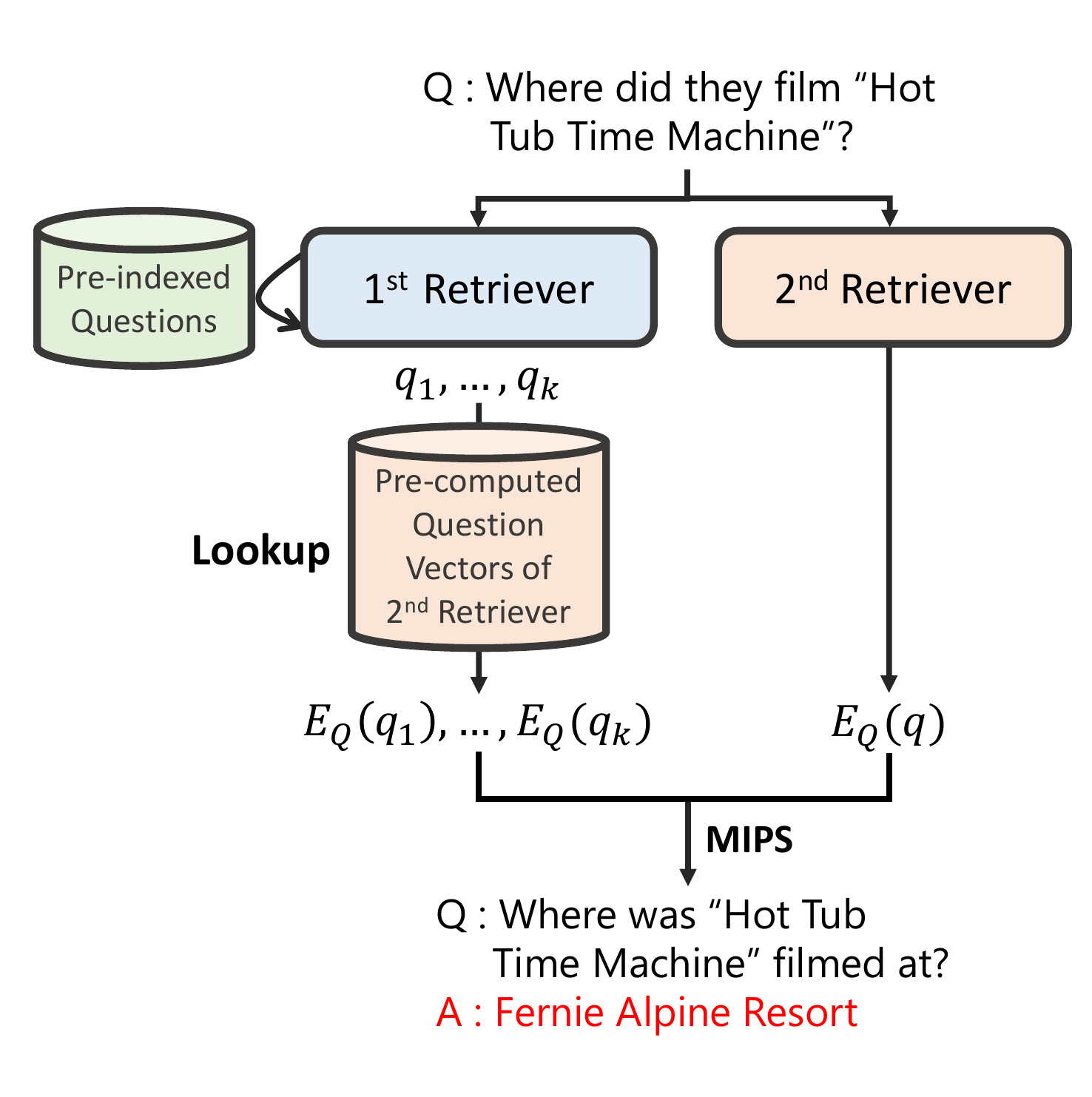}
            \vspace{-2.0em}
            \caption{Inference procedure}
            \label{model_inference}
        \end{subfigure}\hfill%
        % \hspace{.001\textwidth}
    \end{center}
    \vspace{-1.0em}
    \caption{
    \yeon{Illustrations of training and inference processes of \modelname{}.}
    \modelname{} consists of two retrievers. The first-step retriever selects top-k similar questions among the pre-indexed QAs.
    From the top-k results,
    (a) the second-step retriever is trained to distinguish the positive sample from the negative samples, and
    (b) it selects the most similar question at the inference time.
    }
    \label{fig:model}
    \vspace{-0.50em}
\end{figure*}
Our method is constructed based on the question retrieval pipeline proposed by \newcite{lewis2021paq}, where question retrievers find the most similar question to the input question and return the answer of the selected question.
In this study, we note that previous question retrievers are optimized not just for improving the retrieval performance but for maintaining the inference speed to cover millions of text \cite{lewis2021paq}.
In this process, the performance of retrievers decreases as they are more optimized for computational efficiency.
We propose to use an additional retriever that takes the top-k predictions from the first retriever and selects the most similar question from the top-k results.
The second-step retriever has a lower constraint in the inference speed than the first retriever since its search space contains only a few samples.
This enables us to focus only on the retrieval performance when designing the training method.
The overall training and inference procedure of \modelname{} is illustrated in Figure \ref{fig:model}.
We describe the details of \modelname{} below.

\subsection{Training}
Since the annotated question-question pairs are unavailable, we distantly supervise \modelname{} with heuristically selected positive and negative samples.
We first select top-k similar questions with the first-step retriever.
Among the top-k questions, we choose the positive samples and the negative samples as the following.
For positive samples, we choose questions with the most similar answer to the ground truth answer in terms of F1-score, the evaluation metric used in extractive QA \cite{rajpurkar2016squad}. 
For negative samples, we sample questions with answers that differ from the ground truth answer \cite{karpukhin2020dense, xiong2020approximate}.

When the input question is provided with a positive sample ($q^+$) and $m$ negative samples %($\{q_1^-,...,q_m^-\}$),
($q_1^-,...,q_m^-$), our second-step retriever is trained to distinguish the positive and negative samples. The loss function is as follows:
\begin{equation}
    \begin{split}
        L(q, q^+, q_1^-,& ..., q_m^-) = \\ -\log&(\frac{e^{\text{sim}(q, q^+)}}{e^{\text{sim}(q, q^+)} + \sum_{i=1}^m {e^{\text{sim}(q, q_i^-)}}}).
    \end{split}
\end{equation}
The similarity function is defined as the dot product of two vectors:
$\text{sim}(q_1, q_2) = E_Q(q_1)^TE_Q(q_2)$.
%. \begin{equation}
%     \text{sim}(q_1, q_2) = E_Q(q_1)^TE_Q(q_2),    
% \end{equation}
Where $E_Q(·)$ is the question encoder of the second-step retriever.

\subsection{Inference}
Given a question $q$, the two retrievers of \modelname{} work in two steps.
The first-step retriever selects top-k similar questions.
The retrieved questions are then mapped to the question vectors pre-computed by the second-step retriever.
The second-step retriever selects the most similar question $q^{\prime}$ from the top-k results with the question vectors.
We use Maximum Inner Product Search (MIPS) for the second-step retrieval.
Finally, \modelname{} puts the answer of $q^{\prime}$ as the answer for $q$.
\section{Experimental Setup and Results}
\begin{table*}[]
\centering
\begin{tabular}{@{}llccc@{}}
\toprule
Model Type                          & Model                                           & \multicolumn{1}{l}{NQ} & \multicolumn{1}{l}{TriviaQA} & \multicolumn{1}{l}{Inference speed (Q/sec)} \\ \midrule
\multirow{5}{*}{Question retrieval} & RePAQ-base256 \cite{lewis2021paq}               & 40.0                   & 38.8                         & 1376                                        \\
                                    & RePAQ-base \cite{lewis2021paq}                  & 40.9                   & 39.7                         & 738                                         \\
                                    & RePAQ-large \cite{lewis2021paq}                 & 41.2                   & 38.8                         & 624                                         \\
                                    & \modelname{}-BM25/DPR                           & 43.1                   & \textbf{45.6}                & 328                                         \\
                                    & \modelname{}-RePAQ/DPR                          & \textbf{44.0}          & 44.9                         & 1006 (1266\SUP{$\dagger$})                  \\ \midrule
Phrase-indexed                      & DensePhrase \cite{lee-etal-2021-learning-dense} & 40.9                   & 50.7                         & 20.6\SUP{*}                                 \\ \midrule
\multirow{2}{*}{Retriever-reader}   & RAG-Sequence \cite{NEURIPS2020_6b493230}        & 44.5                   & 56.8                         & 0.8                                         \\
                                    & FiD-large \cite{izacard2021leveraging}          & 51.4                   & 67.6                         & 0.5\SUP{*}                                  \\ \bottomrule
\end{tabular}
\caption{
The open-domain QA performance (EM) and inference speeds of \modelname{} and baselines on NQ test set and TriviaQA test set.
We use the performance and the inference speed of each baseline reported from their results.}
* indicates the inference speed is from the original paper.
\SUP{$\dagger$} indicates that the inference speed is computed in the parallel computing setting.
% \vspace{-1em}
\label{tab:odqa}
\end{table*}

\begin{table}[]
\centering
\begin{tabular}{@{}lcc@{}}
\toprule
Supervision      & \multicolumn{1}{l}{BM25} & \multicolumn{1}{l}{RePAQ} \\ \midrule
w/o 2nd retriever & 34.4                     & 40.0                     \\ \midrule
+ Self           & 39.5                     & 40.4                      \\
+ Similar         & 43.1                     & 44.0                      \\
+ Similar / Self  & 43.6                     & 44.1                      \\
+ Same Answer    & 43.4                     & 44.4                      \\ \bottomrule
\end{tabular}
\caption{The open-domain QA performance (EM) of \modelname{} in four different distant supervision methods on NQ test set.}
\vspace{-1.5em}
\label{tab:distant_supervision}
\end{table}

We evaluate the performance and computational efficiency of \modelname{} on two open-domain QA datasets: NaturalQuestions (NQ) and TriviaQA.
We also compare various distant supervision methods for training \modelname{}.
We use exact match (EM) \cite{rajpurkar2016squad} for performance evaluation and the number of questions per second (Q/sec) for evaluation of inference speed.
The details of our experimental setup is described in Appendix \ref{sec:appendix_exp_setup}.

\paragraph{Question Retrievers on Open-Domain QA:}
We evaluate \modelname{} with two different first-step retrievers: BM25 and RePAQ\jiho{-base256}\footnote{We use RePAQ-base256 provided by the official implementation. RePAQ-base256 has slightly lower performance than RePAQ-base.} \cite{lewis2021paq}.
Table \ref{tab:odqa} shows that \modelname{}-BM25/DPR and \modelname{}-RePAQ/DPR achieve the best performance among question retrieval models on TriviaQA and NQ, respectively.
Note that \modelname{}-RePAQ/DPR outperforms RePAQ-base256 significantly with a negligible loss of inference speed; 4.0\%p EM gain on NQ and 6.1\%p gain on TriviaQA at 92.0\% speed (1266 Q/sec vs. 1376 Q/sec). 

\paragraph{Trade-off between QA Performance and Computational Efficiency:}
Table \ref{tab:odqa} shows the trade-off between the open-domain QA performance and the inference speed of the three types of open-domain QA models.
Comparing RePAQ-large and RAG-Sequence, we see a large performance gap of 3.3\%p on NQ and 18.0\%p on TriviaQA, and we also see a large speed gap of 624 Q/s and 0.8 Q/s.
\modelname{} bridges this gap, achieving comparable performances to RAG-Sequence on NQ while maintaining the high inference speed. The performance gain on TriviaQA is not as high, and we conjecture that this is because RePAQ uses only questions from NQ in its filtering step. We leave a deeper study of this discrepancy for future research.

Figure \ref{fig:comp_eff} illustrates the QA performance and inference speed of various configurations of RePAQ \modelname{}.
We vary the encoder of the second-step retriever with different pre-trained models: DPR \cite{karpukhin2020dense}, BERT-base/large \cite{devlin2019bert}, and ALBERT-base/large \cite{lan2019albert}.
The first and second-step question encoders can be executed concurrently, so we run them in parallel and set the batch size as half to measure the inference speed (\modelname{}-DPR-parallel).
We use the maximum batch size possible on a single V100-16GB GPU.
The figure shows that results of \modelname{} all lie to the top right of the curve fitted to the RePAQ results, meaning that \modelname{} succeeds in improving both QA performance and inference speed. The detailed results are in Appendix \ref{sec:appendix_tradeoff}.

\paragraph{Analysis on Positive Sampling Methods:}
We distantly supervise the second-step retriever because annotated question-question pairs are unavailable.
We conduct experiments on various positive sampling methods for distant supervision: ``Self'', ``Similar'', ``Similar/Self'', and ``Same Answer''.
Each method uses the following as the positive sample:

1) the input question itself (``Self''),
2) a similar question with a similar answer (``Similar''),
3) a similar question if it has the ground truth answer, or the input question itself (``Similar/Self''), and
4) a random question with the ground truth answer (``Same Answer'').

Table \ref{tab:distant_supervision} shows the performance of \modelname{}-BM25 and \modelname{}-RePAQ-base256 on the NQ test set with the four distant supervision methods.
The first row (w/o 2nd retriever) indicates the performance based only on the first-step retriever (BM25 or RePAQ-base256).
The second-step retriever with ``Self'' method improves the performance slightly, and the others improve the performance more significantly.
The large gap between ``Self'' and the other methods shows that using the answer information is essential for distant supervision.

\paragraph{Error Propagation Analysis:}
The error rate of each stage in a multi-stage model provides a better understanding of the model's performance boundary.
In \modelname{}, the second-step retriever only predicts the correct answer when the top-50 question-answer pairs retrieved by the first-step retriever contain the answer.
This indicates that the upper-bound performance of \modelname{} is determined by the performance of the first-step retriever.
We measure the R@50 accuracy of the first-step retrievers on NQ and TriviaQA.
The performance of BM25 and RePAQ are 64.07\% and 64.34\% on NQ and 61.73\% and 59.10\% on TriviaQA, respectively.
\section{Conclusion}
The trade-off between the performance and the inference speed is an important problem in open-domain QA.
Recently proposed question retrieval models have shown significantly improved inference speed.
However, this improvement came at the cost of a significantly lower QA performance by the question retrieval models compared to the state-of-the-art open-domain QA models.
In this paper, we proposed a two-step question retrieval model, \modelname{}.
We evaluated the open-domain QA performance and the inference speed of \modelname{} on two datasets: NaturalQuestions and TriviaQA.
From the results, we showed that the sequential two-retriever approach in \modelname{} achieves a significant QA performance improvement over the existing question retrieval models, while retaining the advantage of faster inference speed. This improvement in both QA performance and inference speed is a meaningful step toward the development of real-time open domain QA systems.
\section*{Acknowledgements}
This work was partly supported by NAVER Corp. and the Engineering Research Center Program through the National Research Foundation of Korea (NRF) funded by the Korean Government MSIT (NRF-2018R1A5A1059921).

% Entries for the entire Anthology, followed by custom entries
\bibliography{acl}

\begin{thebibliography}{26}
\expandafter\ifx\csname natexlab\endcsname\relax\def\natexlab#1{#1}\fi

\bibitem[{Devlin et~al.(2019)Devlin, Chang, Lee, and
  Toutanova}]{devlin2019bert}
Jacob Devlin, Ming-Wei Chang, Kenton Lee, and Kristina Toutanova. 2019.
\newblock {BERT}: Pre-training of deep bidirectional transformers for language
  understanding.
\newblock In \emph{NAACL-HLT}.

\bibitem[{Du et~al.(2017)Du, Shao, and Cardie}]{du2017learning}
Xinya Du, Junru Shao, and Claire Cardie. 2017.
\newblock Learning to ask: Neural question generation for reading
  comprehension.
\newblock In \emph{ACL}.

\bibitem[{Duan et~al.(2017)Duan, Tang, Chen, and Zhou}]{duan2017question}
Nan Duan, Duyu Tang, Peng Chen, and Ming Zhou. 2017.
\newblock Question generation for question answering.
\newblock In \emph{EMNLP}.

\bibitem[{Fabbri et~al.(2020)Fabbri, Ng, Wang, Nallapati, and
  Xiang}]{fabbri2020template}
Alexander~Richard Fabbri, Patrick Ng, Zhiguo Wang, Ramesh Nallapati, and Bing
  Xiang. 2020.
\newblock Template-based question generation from retrieved sentences for
  improved unsupervised question answering.
\newblock In \emph{ACL}.

\bibitem[{Izacard and Grave(2021)}]{izacard2021leveraging}
Gautier Izacard and {\'E}douard Grave. 2021.
\newblock Leveraging passage retrieval with generative models for open domain
  question answering.
\newblock In \emph{EACL}.

\bibitem[{Joshi et~al.(2017)Joshi, Choi, Weld, and
  Zettlemoyer}]{joshi2017triviaqa}
Mandar Joshi, Eunsol Choi, Daniel~S Weld, and Luke Zettlemoyer. 2017.
\newblock {TriviaQA}: A large scale distantly supervised challenge dataset for
  reading comprehension.
\newblock In \emph{ACL}.

\bibitem[{Karpukhin et~al.(2020)Karpukhin, Oguz, Min, Lewis, Wu, Edunov, Chen,
  and Yih}]{karpukhin2020dense}
Vladimir Karpukhin, Barlas Oguz, Sewon Min, Patrick Lewis, Ledell Wu, Sergey
  Edunov, Danqi Chen, and Wen-tau Yih. 2020.
\newblock Dense passage retrieval for open-domain question answering.
\newblock In \emph{EMNLP}.

\bibitem[{Kwiatkowski et~al.(2019)Kwiatkowski, Palomaki, Redfield, Collins,
  Parikh, Alberti, Epstein, Polosukhin, Devlin, Lee
  et~al.}]{kwiatkowski2019natural}
Tom Kwiatkowski, Jennimaria Palomaki, Olivia Redfield, Michael Collins, Ankur
  Parikh, Chris Alberti, Danielle Epstein, Illia Polosukhin, Jacob Devlin,
  Kenton Lee, et~al. 2019.
\newblock Natural questions: A benchmark for question answering research.
\newblock \emph{TACL}.

\bibitem[{Lan et~al.(2019)Lan, Chen, Goodman, Gimpel, Sharma, and
  Soricut}]{lan2019albert}
Zhenzhong Lan, Mingda Chen, Sebastian Goodman, Kevin Gimpel, Piyush Sharma, and
  Radu Soricut. 2019.
\newblock {ALBERT}: A lite bert for self-supervised learning of language
  representations.
\newblock In \emph{ICLR}.

\bibitem[{Lee et~al.(2020)Lee, Seo, Hajishirzi, and
  Kang}]{lee2020contextualized}
Jinhyuk Lee, Minjoon Seo, Hannaneh Hajishirzi, and Jaewoo Kang. 2020.
\newblock Contextualized sparse representations for real-time open-domain
  question answering.
\newblock In \emph{ACL}.

\bibitem[{Lee et~al.(2021{\natexlab{a}})Lee, Sung, Kang, and
  Chen}]{lee-etal-2021-learning-dense}
Jinhyuk Lee, Mujeen Sung, Jaewoo Kang, and Danqi Chen. 2021{\natexlab{a}}.
\newblock Learning dense representations of phrases at scale.
\newblock In \emph{ACL}.

\bibitem[{Lee et~al.(2021{\natexlab{b}})Lee, Wettig, and Chen}]{lee2021phrase}
Jinhyuk Lee, Alexander Wettig, and Danqi Chen. 2021{\natexlab{b}}.
\newblock Phrase retrieval learns passage retrieval, too.
\newblock \emph{arXiv}.

\bibitem[{Lewis et~al.(2020{\natexlab{a}})Lewis, Liu, Goyal, Ghazvininejad,
  Mohamed, Levy, Stoyanov, and Zettlemoyer}]{lewis2020bart}
Mike Lewis, Yinhan Liu, Naman Goyal, Marjan Ghazvininejad, Abdelrahman Mohamed,
  Omer Levy, Veselin Stoyanov, and Luke Zettlemoyer. 2020{\natexlab{a}}.
\newblock {BART}: Denoising sequence-to-sequence pre-training for natural
  language generation, translation, and comprehension.
\newblock In \emph{ACL}.

\bibitem[{Lewis et~al.(2020{\natexlab{b}})Lewis, Perez, Piktus, Petroni,
  Karpukhin, Goyal, K\"{u}ttler, Lewis, Yih, Rockt\"{a}schel, Riedel, and
  Kiela}]{NEURIPS2020_6b493230}
Patrick Lewis, Ethan Perez, Aleksandra Piktus, Fabio Petroni, Vladimir
  Karpukhin, Naman Goyal, Heinrich K\"{u}ttler, Mike Lewis, Wen-tau Yih, Tim
  Rockt\"{a}schel, Sebastian Riedel, and Douwe Kiela. 2020{\natexlab{b}}.
\newblock Retrieval-augmented generation for knowledge-intensive nlp tasks.
\newblock In \emph{NeurIPS}.

\bibitem[{Lewis et~al.(2021{\natexlab{a}})Lewis, Stenetorp, and
  Riedel}]{lewis2021question}
Patrick Lewis, Pontus Stenetorp, and Sebastian Riedel. 2021{\natexlab{a}}.
\newblock Question and answer test-train overlap in open-domain question
  answering datasets.
\newblock In \emph{EACL}.

\bibitem[{Lewis et~al.(2021{\natexlab{b}})Lewis, Wu, Liu, Minervini,
  K{\"u}ttler, Piktus, Stenetorp, and Riedel}]{lewis2021paq}
Patrick Lewis, Yuxiang Wu, Linqing Liu, Pasquale Minervini, Heinrich
  K{\"u}ttler, Aleksandra Piktus, Pontus Stenetorp, and Sebastian Riedel.
  2021{\natexlab{b}}.
\newblock {PAQ}: 65 million probably-asked questions and what you can do with
  them.
\newblock \emph{arXiv}.

\bibitem[{Mao et~al.(2021{\natexlab{a}})Mao, He, Liu, Shen, Gao, Han, and
  Chen}]{mao-etal-2021-generation}
Yuning Mao, Pengcheng He, Xiaodong Liu, Yelong Shen, Jianfeng Gao, Jiawei Han,
  and Weizhu Chen. 2021{\natexlab{a}}.
\newblock Generation-augmented retrieval for open-domain question answering.
\newblock In \emph{ACL}.

\bibitem[{Mao et~al.(2021{\natexlab{b}})Mao, He, Liu, Shen, Gao, Han, and
  Chen}]{mao-etal-2021-reader}
Yuning Mao, Pengcheng He, Xiaodong Liu, Yelong Shen, Jianfeng Gao, Jiawei Han,
  and Weizhu Chen. 2021{\natexlab{b}}.
\newblock Reader-guided passage reranking for open-domain question answering.
\newblock In \emph{ACL-Findings}.

\bibitem[{Rajpurkar et~al.(2016)Rajpurkar, Zhang, Lopyrev, and
  Liang}]{rajpurkar2016squad}
Pranav Rajpurkar, Jian Zhang, Konstantin Lopyrev, and Percy Liang. 2016.
\newblock Squad: 100,000+ questions for machine comprehension of text.
\newblock In \emph{EMNLP}.

\bibitem[{Sachan et~al.(2021)Sachan, Reddy, Hamilton, Dyer, and
  Yogatama}]{sachan2021end}
Devendra~Singh Sachan, Siva Reddy, William Hamilton, Chris Dyer, and Dani
  Yogatama. 2021.
\newblock End-to-end training of multi-document reader and retriever for
  open-domain question answering.
\newblock \emph{arXiv}.

\bibitem[{Seo et~al.(2018)Seo, Kwiatkowski, Parikh, Farhadi, and
  Hajishirzi}]{seo2018phrase}
Minjoon Seo, Tom Kwiatkowski, Ankur Parikh, Ali Farhadi, and Hannaneh
  Hajishirzi. 2018.
\newblock Phrase-indexed question answering: A new challenge for scalable
  document comprehension.
\newblock In \emph{EMNLP}.

\bibitem[{Seo et~al.(2019)Seo, Lee, Kwiatkowski, Parikh, Farhadi, and
  Hajishirzi}]{seo2019real}
Minjoon Seo, Jinhyuk Lee, Tom Kwiatkowski, Ankur Parikh, Ali Farhadi, and
  Hannaneh Hajishirzi. 2019.
\newblock Real-time open-domain question answering with dense-sparse phrase
  index.
\newblock In \emph{ACL}.

\bibitem[{Xiong et~al.(2021)Xiong, Xiong, Li, Tang, Liu, Bennett, Ahmed, and
  Overwijk}]{xiong2020approximate}
Lee Xiong, Chenyan Xiong, Ye~Li, Kwok-Fung Tang, Jialin Liu, Paul~N Bennett,
  Junaid Ahmed, and Arnold Overwijk. 2021.
\newblock Approximate nearest neighbor negative contrastive learning for dense
  text retrieval.
\newblock In \emph{ICLR}.

\bibitem[{Xiong et~al.(2020)Xiong, Li, Iyer, Du, Lewis, Wang, Mehdad, Yih,
  Riedel, Kiela et~al.}]{xiong2020answering}
Wenhan Xiong, Xiang Li, Srini Iyer, Jingfei Du, Patrick Lewis, William~Yang
  Wang, Yashar Mehdad, Scott Yih, Sebastian Riedel, Douwe Kiela, et~al. 2020.
\newblock Answering complex open-domain questions with multi-hop dense
  retrieval.
\newblock In \emph{ICLR}.

\bibitem[{Yamada et~al.(2021)Yamada, Asai, and Hajishirzi}]{yamada2021bpr}
Ikuya Yamada, Akari Asai, and Hannaneh Hajishirzi. 2021.
\newblock Efficient passage retrieval with hashing for open-domain question
  answering.
\newblock In \emph{ACL}.

\bibitem[{Zhao et~al.(2021)Zhao, Lu, and Lee}]{zhao-etal-2021-sparta}
Tiancheng Zhao, Xiaopeng Lu, and Kyusong Lee. 2021.
\newblock {SPARTA}: Efficient open-domain question answering via sparse
  transformer matching retrieval.
\newblock In \emph{NAACL-HLT}.

\end{thebibliography}
\bibliographystyle{acl_natbib}

\appendix
\section{Appendix}
\subsection{Detailed results of Figure 1}\label{sec:appendix_tradeoff}
Table \ref{tab:appendix_tradeoff} shows the detailed results of Figure \ref{fig:comp_eff}.

\begin{table}
\centering
\begin{tabular}{@{}lcc@{}}
\toprule
Model      & \multicolumn{1}{l}{EM} & \multicolumn{1}{l}{Q/sec} \\ \midrule
\modelname{}-RePAQ/DPR-parallel & \textbf{44.0}                     & \textbf{1266}                     \\ 
\modelname{}-RePAQ/DPR & \textbf{44.0}                     & 1006                     \\ 
\modelname{}-RePAQ/BERT-large     & 43.1                     & 814          \\
\modelname{}-RePAQ/BERT-base     & 43.1                     & 1006          \\
\modelname{}-RePAQ/ALBERT-large     & 42.2                     & 677          \\
\modelname{}-RePAQ/ALBERT-base     & 41.8                    & 920          \\ \midrule
RePAQ-base256         & 40.0                    & \textbf{1376}                      \\
RePAQ-large         & 41.2                     & 624                      \\
RePAQ-xlarge         & 41.5                     & 467                      \\
RePAQ-base + Reranker-base         & 45.7                     & 41        \\
RePAQ-large + Reranker-xlarge         & \textbf{46.2}                     & 7        \\ \bottomrule
\end{tabular}
\caption{EM score and inference speed on NQ for various configurations of \modelname{} and RePAQ}
\label{tab:appendix_tradeoff}
\end{table}

\subsection{Experimental Setup}\label{sec:appendix_exp_setup}
\paragraph{Training Details:}
\yeon{We set the batch size to 2 per GPU and the number of negative samples to 16.
We used validation EM score for early stopping.
\modelname{} was trained on a machine with four V100-16GB GPUs.
We report the result of a single trial.
}
\paragraph{Computational Environment for Measuring the Inference Speed:}
The inference speed of baseline models and \modelname{} is measured with a V100-16GB GPU and 32 CPUs (Intel Xeon E5-2686v4).
\yeon{We report mean of three separate trials.}
\subsection{License or Terms of Artifacts}
We use BERT whose license is under the Apache License 2.0 free with modification and distribution. Also, we use RePAQ whose license is under the CC BY-NC 4.0 free with modification and distribution. All models we used are publicly available.

\end{document}